\documentclass[11pt,a4paper]{article}
\usepackage[hyperref]{acl2021}
\usepackage{times}
\usepackage{latexsym}

\usepackage{microtype}

\aclfinalcopy 
\usepackage{graphicx}
\usepackage{amsmath}
\usepackage{amssymb}
\usepackage{amsfonts}
\usepackage{amstext}
\usepackage{amsthm}
\usepackage{bbm}
\usepackage{arydshln}
\usepackage{tikz}
\usepackage{tikz-dependency}
\usepackage{graphicx}
\usepackage{multirow}
\usepackage{tikz-qtree}
\usepackage{algorithm}
\usepackage[T1]{fontenc}
\usepackage[tableaux]{prooftrees}
\usepackage{qtree}

\usepackage{caption}
\usepackage{subcaption}
\usepackage[export]{adjustbox}
\usepackage{algorithm}
\usepackage{colortbl}
\usepackage{tabstackengine}
\usepackage{arydshln}
\usepackage[colorinlistoftodos]{todonotes}

\pgfdeclarelayer{background}
\pgfdeclarelayer{foreground}
\pgfsetlayers{background,main,foreground}

\usepackage[noend]{algpseudocode}
\algnewcommand{\parState}[1]{\State%
    \parbox[t]{\dimexpr\linewidth-\algmargin}{\strut\hangindent=\algorithmicindent \hangafter=1 #1\strut}}

\algrenewcommand\algorithmicindent{1.0em}%

\DeclareMathOperator*{\argmax}{arg\,max}
\DeclareMathOperator*{\argmin}{arg\,min}

\newcommand{\comment}[1]{}

\newcommand{\yy}{\mathbf{y}}

\newcommand{\calH}{\mathcal{H}}
\newcommand{\calP}{\mathcal{P}}
\newcommand{\calS}{\mathcal{S}}

\newcommand{\calQ}{\mathcal{Q}}

\newcommand{\score}{\mathrm{score}}

\tikzset{
var node/.style={circle,draw,node distance=2.0cm}, 
exvar node/.style={dotted,circle,draw,node distance=2.0cm,thick}, 
pred node/.style={rectangle,rounded corners,draw,node distance=1.7cm}, 
rel node/.style={fill=white,circle,inner sep=0.1mm}
}

\newcommand{\redb}{red!55}

\usepackage[symbol]{footmisc}
\title{NeuralLog: \\Natural Language Inference with Joint Neural and Logical Reasoning}

\author{
{Zeming Chen$^{\dagger}$\footnotemark} \quad
  Qiyue Gao$^{\dagger}$ \quad
  \textbf{Lawrence S. Moss}$^\ddagger$ \\
$^\dagger$Rose-Hulman Institute of Technology, Terre Haute, IN, USA\\
$^\ddagger$Indiana University, Bloomington, IN, USA\\
{\tt{\{chenz16,gaoq\}@rose-hulman.edu}} \\
{\tt \{lmoss\}@indiana.edu}
}

\date{}

\begin{document}
\maketitle
\footnotetext[1]{The first two authors have equal contribution}
\begin{abstract}
Deep learning (DL) based language models achieve high performance on various benchmarks for Natural Language Inference (NLI). And at this time, symbolic approaches to NLI are receiving less attention. Both approaches (symbolic and DL) have their advantages and weaknesses. However, currently, no method combines them in a system to solve the task of NLI. To merge symbolic and deep learning methods, we propose an inference framework called NeuralLog, which utilizes both a monotonicity-based logical inference engine and a neural network language model for phrase alignment. Our framework models the NLI task as a classic search problem and uses the beam search algorithm to search for optimal inference paths. Experiments show that our joint logic and neural inference system improves accuracy on the NLI task and can achieve state-of-art accuracy on the SICK and MED datasets.  
\end{abstract}

\section{Introduction}
 \begin{figure}[t!]
    \centering
    \includegraphics[width=5.5cm]{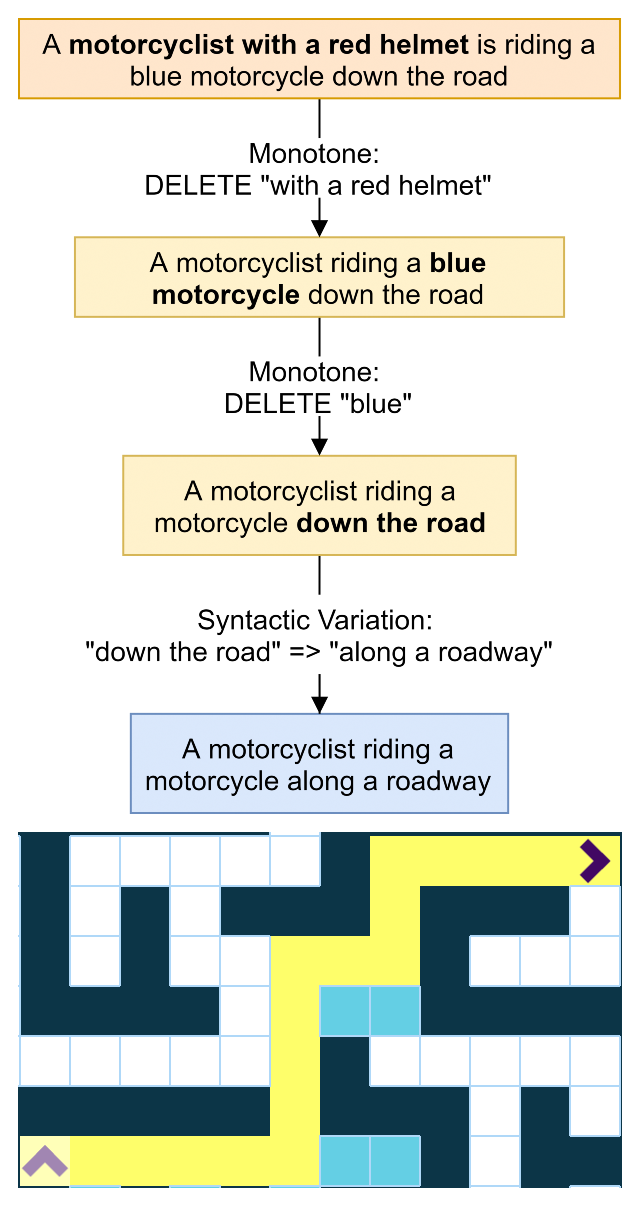}
    \caption{Analogy between path planning and an entailment inference path from the premise \textit{A motorcyclist with a red helmet is riding a blue motorcycle down the road} to the hypothesis \textit{A motorcyclist is riding a motorbike along a roadway}.}
    \label{fig:proof_path}
\end{figure}
Currently, many NLI benchmarks’ state-of-the-art systems are exclusively deep learning (DL) based language models \cite{devlin-etal-2019-bert, Lan2020ALBERT:, liu2020roberta, yin-schutze-2017-task}. These models often contain a large number of parameters, use high-quality pre-trained embeddings, and are trained on large-scale datasets, which enable them to handle diverse and large test data robustly. However,  several experiments show that DL models lack generalization ability, adopt fallible syntactic heuristics, and show exploitation of annotation artifacts \cite{glockner-etal-2018-breaking, mccoy-etal-2019-right, gururangan-etal-2018-annotation}. On the other hand, there are logic-based systems that use symbolic reasoning and semantic formalism to solve NLI \cite{abzianidze-2017-langpro, martinez-gomez-etal-2017-demand, yanaka-etal-2018-acquisition, hu-etal-2020-monalog}. These systems show high precision on complex inferences involving difficult linguistic phenomena and present logical and explainable reasoning processes. However, these systems lack background knowledge and do not handle sentences with syntactic variations well, which makes them poor competitors with state-of-the-art DL models. Both DL and logic-based systems show a major issue with NLI models: they are too one-dimensional (either purely DL or purely logic), and no method has combined these two approaches together for solving NLI. 

This paper makes several contributions, as follows: first, we propose a new framework in section 3 for combining logic-based inference with deep-learning-based network inference for better performance on conducting natural language inference. We model an NLI task as a path-searching problem between the premises and the hypothesis. We use beam-search to find an optimal path that can transform a premise to a hypothesis through a series of inference steps. This way, different inference modules can be inserted into the system. For example, DL inference modules will handle inferences with diverse syntactic changes and logic inference modules will handle inferences that require complex reasoning. Second, we introduce a new method in section 4.3 to handle syntactic variations in natural language through sequence chunking and DL based paraphrase detection. We evaluate our system in section 6 by conducting experiments on the SICK and MED datasets. Experiments show that joint logical and neural reasoning show state-of-art accuracy and recall on these datasets.

\section{Related Work}
Perhaps the closest systems to NeuralLog are
\citet{yanaka-etal-2018-acquisition}, MonaLog \cite{hu-etal-2020-monalog}, and Hy-NLI \cite{kalouli-etal-2020-hy}. Using \citet{ccg2lambda} to work with logic representations derived from CCG trees, \citet{yanaka-etal-2018-acquisition} proposed a framework that can detect phrase correspondences for a sentence pair, using
natural deduction on semantic relations and can thus extract various paraphrases automatically. Their experiments show that assessing phrase correspondences helps improve NLI accuracy. Our system uses a similar methodology to solve syntactic variation inferences, where we determine if two phrases are paraphrases. Our method is rather different on this point, since we call on neural language models to detect paraphrases between two sentences. We feel that it would be interesting to compare the systems on a more theoretical level, but we have not done the comparison in this paper.

NeuralLog inherits the use of polarity marking found in MonaLog~\cite{hu-etal-2020-monalog}.
(However, we use the dependency-based system of \citet{chengaoudep2mono} instead of the CCG-based system of \citet{hu-moss-2018-polarity}.) MonaLog did propose some integration with neural models, using BERT when logic failed to find entailment or contradiction. We are doing something very different, using neural models to detect paraphrases at several levels of ``chunking''. In addition, the exact algorithms found in Sections 3 and 4 are new here. In a sense, our work on alignment in NLI goes back to \citet{MacCartneyManning} where alignment was used to find a chain of edits that changes a premise to a hypothesis, but our work uses much that simply was not available in 2009.

Hy-NLI is a hybrid system that makes inferences using either symbolic or deep learning models based on how linguistically challenging a pair of sentences is. The principle Hy-NLI followed is that deep learning models are better at handling sentences that are linguistically less complex, and symbolic models are better for sentences containing hard linguistic phenomena. Although the system integrates both symbolic and neural methods, its decision process is still separate, in which the symbolic and deep learning sides make decisions without relying on the other side. Differently, our system incorporates logical inferences and neural inferences as part of the decision process, in which the two inference methods rely on each other to make a final decision.  

\begin{figure*}[t!]
    \centering
    \includegraphics[width=0.9\textwidth]{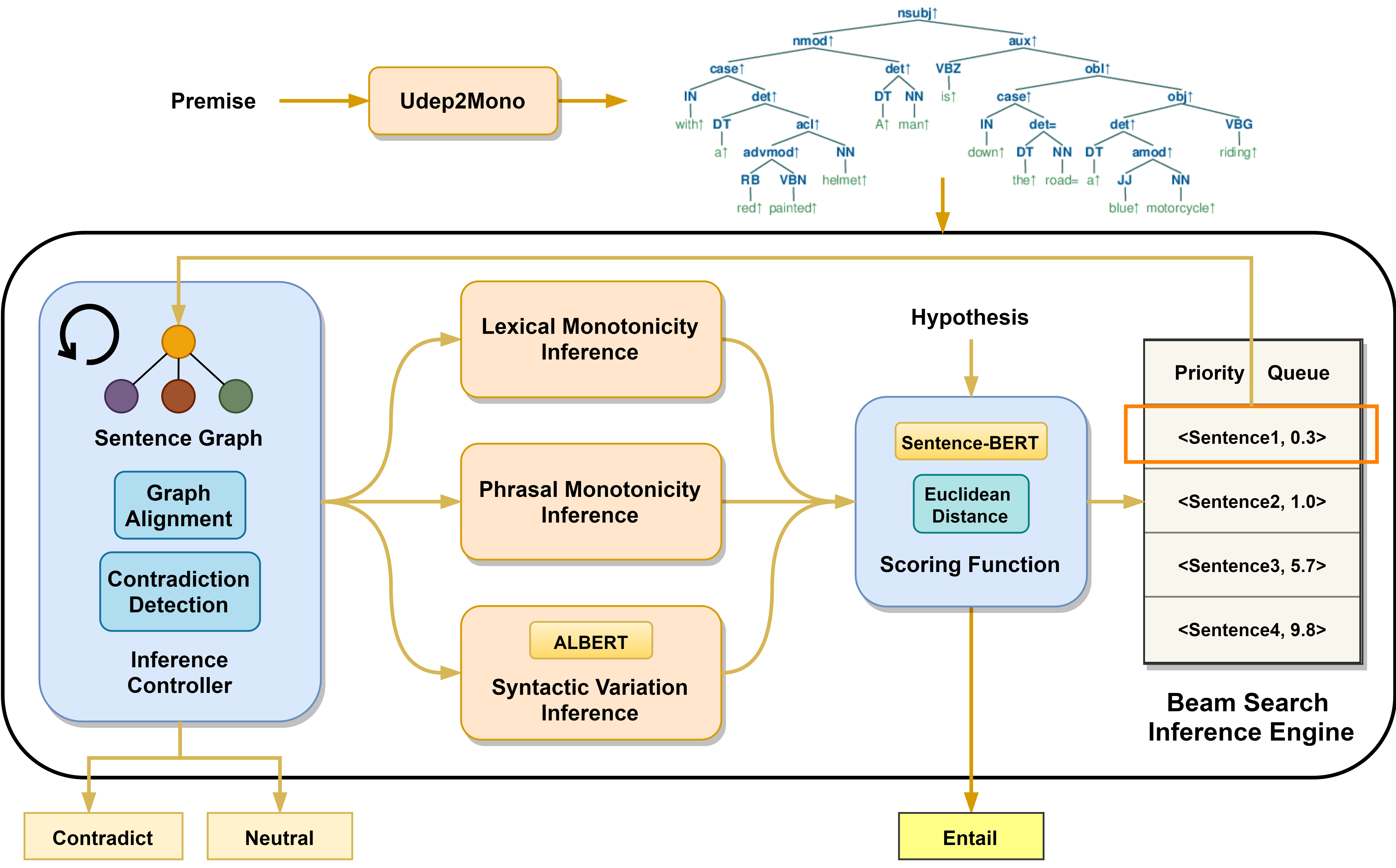}
    \caption{Overview system diagram of NeuralLog.}
    \label{fig:neurallog}
\end{figure*}

 
\section{Method}
\subsection{NLI As Path Planning}
The key motivation behind our architecture and inference modules is that the Natural Language Inference task can be modeled as a path planning problem. Path planning is a task for finding an optimal path traveling from a start point to a goal containing a series of actions. To formulate NLI as path planning, we define the \textbf{premise} as the \textbf{start state} and the \textbf{hypothesis} as the \textbf{goal} that needs to be reached. The classical path planning strategy applies expansions from the start state through some search algorithms, such as depth-first-search or Dijkstra search, until an expansion meets the goal. In a grid map, two types of action produce an expansion. The vertical action moves up and down, and the horizontal action moves left and right.
Similarly, language inference also contains these two actions. Monotonicity reasoning is a vertical action, where the monotone inference moves up and simplifies a sentence, and the antitone inference moves down and makes a sentence more specific. Syntactic variation and synonym replacement are horizontal actions. They change the form of a sentence while maintaining the original meaning. Then, similar to path planning, we can continuously make inferences from the premise using a search algorithm to determine if the premise entails the hypothesis by observing whether one of the inferences can reach the hypothesis. If the hypothesis is reached, we can connect the list of inferences that transform a premise to a hypothesis to be the optimal path in NLI, a valid reasoning chain for entailment.


Figure \ref{fig:proof_path} shows an analogy between an optimal path for the classical grid path planning problem and an example of an optimal inference path for NLI. On the top, we have a reasoning process for natural language inference. From the premise, we can first delete the modifier \textit{with a red helmet}, then delete \textit{blue} to get a simplified sentence. Finally, we can paraphrase \textit{down the road} to \textit{along a roadway} in the premise to reach the hypothesis and conclude the entailment relationship between these two sentences.

\subsection{Overview}
Our system contains four components: (1) a polarity annotator, (2) three sentence inference modules, (3) a search engine, and (4) a sentence inference controller. Figure \ref{fig:neurallog} shows a diagram of the full system. The system first annotates a sentence with monotonicity information (polarity marks) using Udep2Mono \cite{chengaoudep2mono}. The polarity marks include monotone ($\uparrow$), antitone ($\downarrow$), and no monotonicity information (=) polarities. Next, the polarized parse tree is passed to the search engine. A beam search algorithm searches for the optimal inference path from a premise to a hypothesis. The search space is generated from three inference modules: lexical, phrasal, and syntactic variation. Through graph alignment, the sentence inference controller selects a inference module to apply to the premise and produce a set of new premises that potentially form entailment relations with the hypothesis. The system returns \textbf{Entail} if an inference path is found. Otherwise, the controller will determine if the premise and hypothesis form a contradiction by searching for counter example signatures and returns \textbf{Contradict} accordingly. If neither \textbf{Entail} nor \textbf{Contradict} is returned, the system returns \textbf{Neutral}.

\subsection{Polarity Annotator}
The system first annotates a given premise with monotonicity information using Udep2Mono, a polarity annotator that determines polarization of all constituents
from universal dependency trees. The annotator first parses the premise into a binarized universal dependency tree and then conducts polarization by recursively marks polarity on each node . An example can be \textit{Every$^\uparrow$ healthy$^\downarrow$ person$^\downarrow$ plays$^\uparrow$ sports$^\uparrow$}.

\subsection{Search Engine}
To efficiently search for the optimal inference path from a premise $\calP$ to a hypothesis $\calH$, we use a beam search algorithm which has the advantage of reducing search space by focusing on sentences with higher scores. To increase the search efficiency and accuracy, we add an inference controller that can guide the search direction. 

\paragraph{Scoring}
In beam search, a priority queue $\calQ$ maintains the set of generated sentences. A core operation is the determination of the highest-scoring generated sentence for a given input under a learned scoring model. In our case, the maximum score is equivalent to the minimum distance:
\begin{align*}
    \yy^\star &= \argmax_{\mathrm{s} \in \calS} \score(\mathrm{s}, \calH) \\
    \yy^\star &= \argmin_{\mathrm{s} \in \calS} \mathrm{dist}(\mathrm{s},\calH)
\end{align*}
where $\calH$ is the hypothesis and $\mathcal{S}$ is a set of generated sentences produced by the three (lexical, phrasal, syntactic variation) inference modules. We will present more details about these inference modules in section 4. 
We formulate the distance function as the Euclidean distance between the sentence embeddings of the premise and hypothesis. To obtain semantically meaningful sentence embeddings efficiently, we use \citet{reimers-gurevych-2019-sentence}'s language model, Sentence-BERT (SBERT), a modification of the BERT model. It uses siamese and triplet neural network structures to derive sentence embeddings which can be easily compared using distance functions. 

\subsection{Sentence Inference Controller}
In each iteration, the search algorithm expands the search space by generating a set of potential sentences using three inference modules: (1) lexical inference, (2) phrasal inference, and (3) syntactic variation inference. To guide the search engine to select the most applicable module, we designed a inference controller that can recommend 
which of the labels the overall algorithm should proceed with. 
For example, for a premise
\textit{All animals eat food} and a hypothesis \textit{All dogs eat food}, only a lexical inference  of \textit{animals} to  \textit{dogs} would be needed. Then, the controller will apply the lexical inference  to the premise, as we discuss below.

\subsubsection{Sentence Representation Graph}
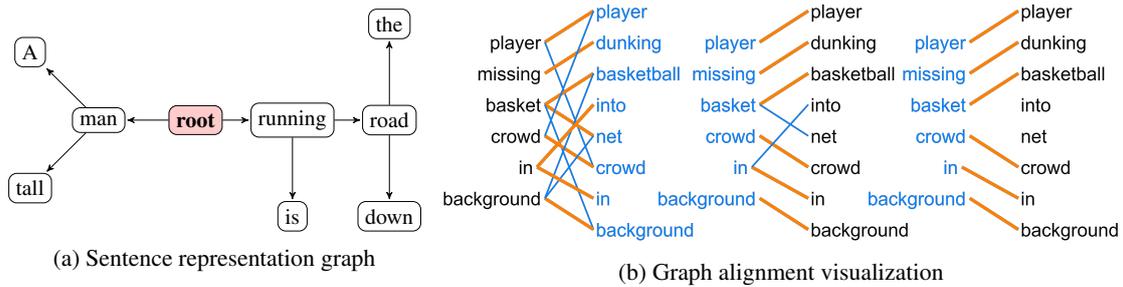
\begin{figure*}[t!]
    \centering
    \begin{subfigure}[h]{0.35\textwidth}
        \centering
        \scalebox{0.75}{
        \begin{tikzpicture}
        [every edge/.style={->,>=stealth',shorten >=1pt,draw},
        caption/.style = {right,inner sep=0mm}]
        
          \node[pred node] (e) [fill=red!20] {\textbf{root}};
          \node[pred node] (man) [left of=e]  {man};
          \node[pred node] (manmod1) [above left of=man]  {A};
          \node[pred node] (manmod2) [below left of=man]  {tall};
          
          \node[pred node] (verb) [right of=e]  {running};
          \node[pred node] (be) [below of=verb]  {is};
          \node[pred node] (verbmod2) [right of=verb]  {road};
          \node[pred node] (verbmod1) [above of=verbmod2]  {the};
          \node[pred node] (verbmod3) [below of=verbmod2]  {down};
          
          \path (e) edge (man);
          \path (e) edge (verb);
          \path (verb) edge (be);
          \path (man) edge (manmod1);
          \path (man) edge (manmod2);
          \path (verbmod2) edge (verbmod1);
          \path (verb) edge (verbmod2);
          \path (verbmod2) edge (verbmod3);
          
        \end{tikzpicture}
        }
        \caption{Sentence representation graph}
        \label{graph}
    \end{subfigure}
    \begin{subfigure}[h]{0.57\textwidth}
        \centering
        \includegraphics[width=\textwidth]{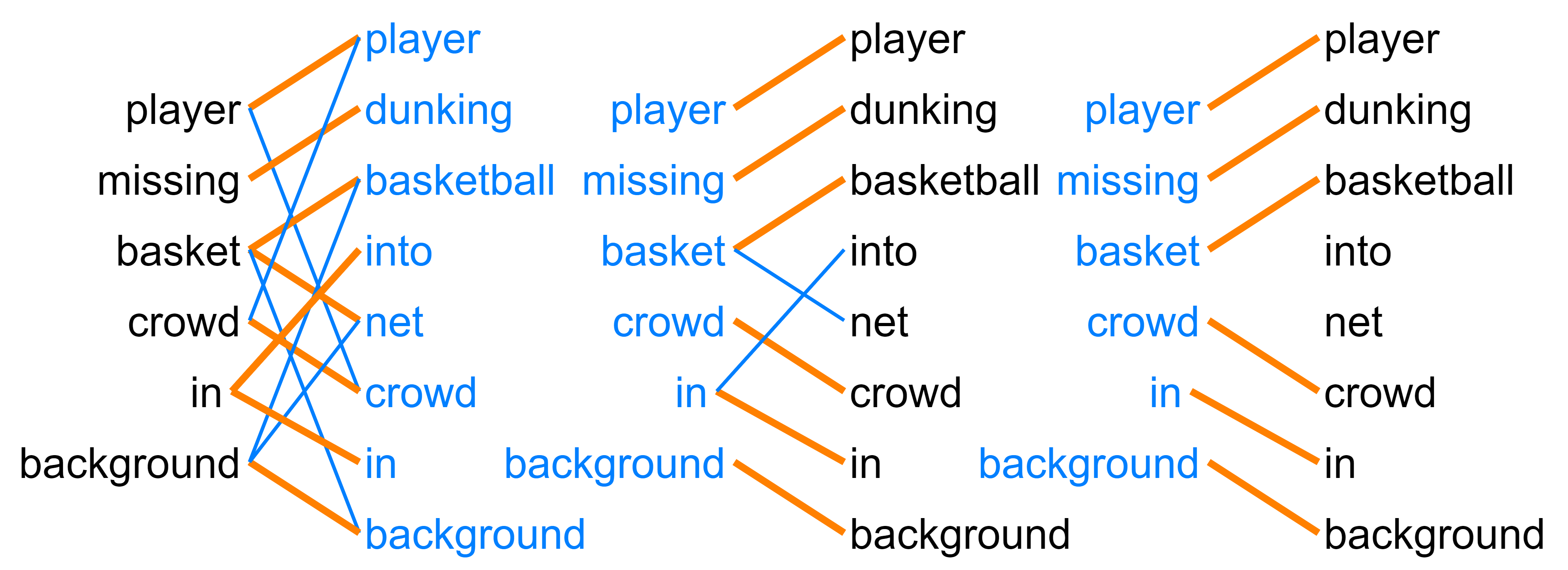}
        \caption{Graph alignment visualization}
        \label{fig:alginment}
    \end{subfigure}
    \caption{(a) A sentence representation graph for \textit{A tall man is running down the road}. (b) Visualization for the graph alignment. The lines between two words represent their similarity. The orange lines are the pairs with maximum similarities for a blue word. Through bi-directional alignment, we eliminate word pairs with non-maximum similarity and gets the final alignment pairs.}
\end{figure*}

The controller makes its decision based on  graph-based representations for the premise
and the hypothesis. We first build a sentence representation graph from
parsed input using Universal Dependencies. Let $\mathcal{V} = \mathcal{V}_{m} \cup \mathcal{V}_{c}$ be the set of vertices of a sentence representation graph, where $\mathcal{V}_{m}$ represents the set of modifiers such as \textit{tall} in Figure \ref{graph-outline}, and ${V}_{c}$ represents the set of content words (words that are being modified) such as \textit{man} in Figure \ref{graph-outline}. While content words in $\mathcal{V}_{c}$ could modify other content words, modifiers in $\mathcal{V}_{m}$ are not modified by other vertices. Let $\mathcal{E}$ be the set of directed edges in the form $\langle v_c, v_m \rangle$ such that $v_m \in \mathcal{V}_{m}$ and $v_c \in \mathcal{V}_{c}$. A sentence representation graph is then defined as a tuple $\mathrm{G} = \langle \mathcal{V}, \mathcal{E} \rangle$. Figure \ref{graph} shows an example graph.

\subsubsection{Graph Alignment}
To observe the differences between two sentences, we rely on graph alignment between two sentence representation graphs. We first align nodes from subjects, verbs and objects, which constitutes what we call a component level. Define $\mathrm{G}_p$ as the graph for a premise and $\mathrm{G}_h$ as the graph for a hypothesis. Then, $\mathcal{C}_p$ and $\mathcal{C}_h$ are component level nodes from the two graphs. We take
the Cartesian product   $\mathcal{C}_p\times \mathcal{C}_h
= \{(\mathrm{c}_p, \mathrm{c}_h): \mathrm{c}_p \in \mathcal{C}_p, \mathrm{c}_h \in \mathcal{C}_h \}$. In the first round, we recursively pair the child nodes of each $\mathrm{c}_p$ to child nodes of each $\mathrm{c}_h$. 
We compute word similarity between two child nodes $\mathrm{c}_p^i$ and $\mathrm{c}_h^i$ and eliminate pairs with non-maximum similarity. We denote the new aligned pairs as a set $\mathcal{A}^*$. At the second round, we iterate through the aligned pairs in $\mathcal{A}^*$. If multiple child nodes from the first graph are paired to a child node in the second graph, we only keep the pair with maximum word similarity. In the final round, we perform the same check for each child node in the first graph to ensure that there are no multiple child nodes from the second graph paired to it. Figure \ref{fig:alginment} shows a brief visualization of the alignment process. 

\subsubsection{inference Module Recommendation}
After aligning the premise graph $\mathcal{G}_{p}$ with hypothesis graph $\mathcal{G}_{h}$, the controller checks through each node in the two graphs. If a node does not get aligned, the controller considers to delete the node or insert it depending on which graph the node belongs to and recommends phrasal inference. If a node is different from its aligned node, the controller recommends lexical inference. If additional lexical or phrasal inferences are detected under this node, the controller decides that there is a more complex transition under this node and recommends a syntactic variation.

\subsubsection{Contradiction Detection}
We determine whether the premise and the hypothesis contradict each other inside the controller by searching for potential contradiction transitions from the premise to the hypothesis. For instance, a transition in the scope of
the quantifier (\textit{a} $\longrightarrow$ \textit{no}) from the same subject could be what we call a contradiction signature
(possible evidence for a contradiction). With all the signatures, the controller decides if they can form a contradiction as a whole. To avoid situations when multiple signatures together fail to form a complete contradiction, such as double negation, the controller checks through the contradiction signatures to ensure a contradiction. For instance, in the verb pair (\textit{not remove}, {\textit{add}}), the contradiction signature \textit{not} would cancel the verb negation contradiction signature from \textit{remove} to \textit{add} so the pair as a whole would not be seen as a contradiction. Nevertheless, other changes from the premise to the hypothesis may change the meaning of the sentence. Hence, our controller would go through other transitions to make sure the meaning of the sentence does not change when the contradiction sign is valid. For example, in the neutral pair P: \textit{A person is eating} and H: \textit{No tall person is eating}, the addition of \textit{tall} would be detected by our controller.
But the aligned word of the component it is applied to, \textit{person} in P, 
has been marked downward monotone.  So this transition is invalid. This pair would then be classified as neutral.

\begin{figure}[h!]
    \centering
    \includegraphics[width=6cm]{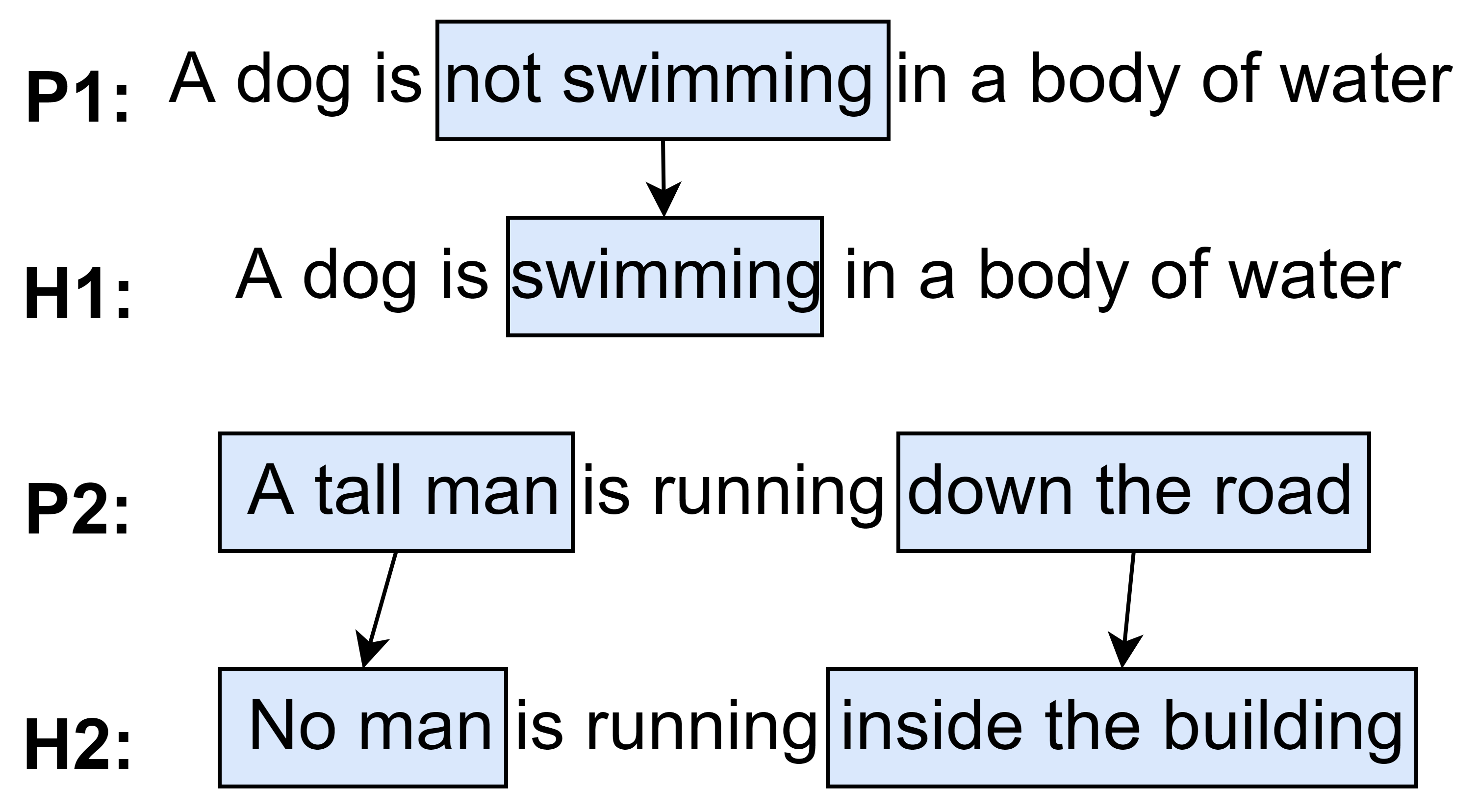}
    \caption{Example of contradiction signatures. P1 and H1 form a contradiction. P2 and H2 does not form a contradiction because the meaning after the verb \textit{running} has changed.}
    \label{fig:contradict}
\end{figure}

\begin{table}[t]
\centering
\scalebox{0.75}{
\centering
\begin{tabular}{ll}
\hline \textbf{signature type} & \textbf{example} \\ \hline
quantifier negation & \textbf{no} dogs $\Longrightarrow$ \textbf{some} dogs\\
verb negation     & is \textbf{eating} $\Longrightarrow$ is \textbf{not eating}\\
noun negation     & \textbf{some people} $\Longrightarrow$ \textbf{nobody} \\
action contradiction & is \textbf{sleeping} $\Longrightarrow$ is \textbf{running} \\
direction contradiction & The \textbf{turtle} is following the \textbf{fish} $\Longrightarrow$ \\
& The \textbf{fish} is following the \textbf{turtle}\\
\hline
\end{tabular}
}
\vspace{-2mm}
\caption{\label{font-table} Examples of contradiction signatures.}
\vspace{-6mm}
\end{table}
For P2 and H2 in Figure \ref{fig:contradict}, the controller notices the contradictory quantifier change around the subject \textit{man}. The subject \textit{man} in P2 is upward monotone so the deletion of \textit{tall} is valid. Our controller also detects the meaning transition from \textit{down the road} to \textit{inside the building}, which affects the sentence's meaning and cancels the previous contradiction signature. The controller thus will not classify P2 and H2 as a pair of contradiction. 

\section{Inference Generation}
\subsection{Lexical Monotonicity Inference} Lexical inference is word replacement based on monotonicity information for key-tokens including nouns, verbs, numbers, and quantifiers. The system uses lexical knowledge bases including WordNet \cite{wordnet} and ConceptNet \cite{conceptnet}. From the knowledge bases, we extract four word sets: hypernyms, hyponyms, synonyms, and antonyms. Logically, if a word has a monotone polarity ($\uparrow$), it can be replaced by its hypernyms. For example, \textit{swim} $\leq$ \textit{move}; then \textit{swim} can be replaced with \textit{move}. If a word has an antitone polarity ($\downarrow$), it can be replaced by its hyponyms. For example, \textit{flower} $\geq$ \textit{rose}. Then, \textit{flower} can be replaced with \textit{rose}. We filter out irrelevant words from the knowledge bases that do not appear in the hypothesis. Additionally, we handcraft knowledge relations for words like quantifiers and prepositions that do not have sufficient taxonomies from knowledge bases. Some handcrafted relations include: \textit{all} = \textit{every} = \textit{each} $\leq$ \textit{most} $\leq$ \textit{many} $\leq$  \textit{several} $\leq$ \textit{some} = \textit{a}, \textit{up} $\perp$ \textit{down}.

\begin{table*}[t!]
\centering
\scalebox{0.85}{
\centering
\begin{tabular}{cll}
\hline \textbf{Type} & \textbf{Premise} & \textbf{Hypothesis} \\ \hline \hline
\multirow{2}{*}{Verb Phrase Variation} & Two men are standing near the water and & Two men are standing near the water and \\ 
& are \textbf{holding fishing poles} & are  \textbf{holding tools used for fishing} \\ \hline
\multirow{2}{*}{Noun Phrase Variation} & A man with climbing equipment is hanging & A man with equipment used for climbing is\\  
& from \textbf{rock which is vertical and white} & hanging from a \textbf{white, vertical rock}. \\ 
\hline
\end{tabular}
}
\caption{\label{font-table} Examples of phrasal alignments detected by the syntactic variation module}
\end{table*}

\begin{figure*}[t]
\small
\centering
\scalebox{0.85}{
\begin{tikzpicture}
[every edge/.style={->,>=stealth',shorten >=1pt,draw},
caption/.style = {right,inner sep=0mm}]
 
  \node[pred node] (e12) [fill=red!20] {\textbf{root}};
  \node[pred node] (man1) [left of=e12]  {man};
  \node[pred node] (man1mod1) [above left of=man1]  {A};
  \node[pred node] (man1mod2) [below left of=man1]  {tall};
  
  \node[pred node] (verb1) [right of=e12]  {running};
  \node[pred node] (bel) [below of=verb1] {is};
  \node[pred node] (verb1mod2) [right of=verb1]  {road};
  \node[pred node] (verb1mod1) [above right of=verb1mod2]  {the};
  \node[pred node] (verb1mod3) [below right of=verb1mod2]  {down};
  
  \path (e12) edge (man1);
  \path (e12) edge (verb1);
  \path (verb1) edge (bel);
  \path (man1) edge (man1mod1);
  \path (man1) [very thick,draw=\redb] edge (man1mod2);
  \path (verb1mod2) [very thick,draw=\redb] edge (verb1mod1);
  \path (verb1) [very thick,draw=\redb] edge (verb1mod2);
  \path (verb1mod2) [very thick,draw=\redb] edge (verb1mod3);
  
  \begin{pgfonlayer}{background}
    \draw[rounded corners=2em,line width=3em,blue!15,cap=round]
     (man1.center) -- (man1mod2.center);
  \end{pgfonlayer}
  
   \begin{pgfonlayer}{background}
    \draw[rounded corners=2em,line width=5em,blue!15,cap=round]
     (verb1.center)  -- (verb1mod2.center) -- (verb1mod1.center) -- (verb1mod2.center) -- (verb1mod3.center) -- (verb1mod2.center);
  \end{pgfonlayer}
  
  \node[pred node] (e1r2) [right=8.5cm of e12,fill=red!20] {\textbf{root}};
  \node[pred node] (man1r) [left of=e1r2]  {man};
  \node[pred node] (manmod1r) [above left of=man1r]  {A};
  \node[pred node] (man1mod1r) [below left of=man1r]  {who};
  \node[pred node] (man1mod2r) [below of=man1r]  {is};
  \node[pred node] (man1mod3r) [below right of=man1r]  {tall};
  
  \node[pred node] (verb1r) [right of=e1r2]  {running};
  \node[pred node] (ber) [below of=verb1r] {is};
  
  \node[pred node] (verb1mod2r) [right of=verb1r]  {roadway};
  \node[pred node] (verb1mod1r) [above right of=verb1mod2r]  {a};
  \node[pred node] (verb1mod3r) [below right of=verb1mod2r]  {along};
  
  \node[pred node] (score1) [above of=verb1, fill=green!50]  {0.98};
  \node[pred node] (score2) [above left of=e1r2, fill=green!50]  {0.99};
  
  \node[pred node] (nscore1) [left of=man1r, fill=yellow!50]  {0.03};
  \node[pred node] (nscore2) [right of=bel, fill=yellow!50]  {0.02};
  
  \path (e1r2) edge (man1r);
  \path (e1r2) edge (verb1r);
  \path (verb1r) edge (ber);
  \path (man1r) edge (manmod1r);
  \path (man1r) [very thick,draw=\redb] [bend right=20] edge (man1mod1r);
  \path (man1r) [very thick,draw=\redb] edge (man1mod2r);
  \path (man1r) [very thick,draw=\redb] [bend left=20] edge (man1mod3r);
  \path (verb1mod2r) [very thick,draw=\redb] edge (verb1mod1r);
  \path (verb1r)[very thick,draw=\redb]  edge (verb1mod2r);
  \path (verb1mod2r) [very thick,draw=\redb] edge (verb1mod3r);
  
  \begin{pgfonlayer}{background}
    \draw[rounded corners=2em,line width=4.5em,blue!15,cap=round]
     (man1r.center) -- (man1mod1r.center) -- (man1mod2r.center) -- (man1mod3r.center) -- (man1r.center);
  \end{pgfonlayer}
  
   \begin{pgfonlayer}{background}
    \draw[rounded corners=2em,line width=5.5em,blue!15,cap=round]
     (verb1r.center)  -- (verb1mod2r.center) -- (verb1mod1r.center) -- (verb1mod2r.center) -- (verb1mod3r.center) -- (verb1mod2r.center);
  \end{pgfonlayer}
  
  \path[dotted] (man1) edge [bend left=15] (score1);
  \path[dotted] (score1) edge [bend left=15] (man1r);
  \path[dotted] (verb1) edge [bend left=15] (score2);
  \path[dotted] (score2) edge [bend left=15] (verb1r);
  
  \path[dotted] (man1) edge [bend right=10] (nscore2);
  \path[dotted] (nscore2) edge [bend right=10] (verb1r);
  \path[dotted] (verb1) edge [bend right=15]  (nscore1);
  \path[dotted] (nscore1) edge (man1r);

\end{tikzpicture}
}

\caption{A graph representation of the monolingual phrase alignment process. Here the left graph represents the premise: \textit{A tall man is running down the road}. The right graph represents the hypothesis \textit{A man who is tall is running along a roadway}. The blue region represents phrase chunks extracted by the chunker from the graph. An alignment score is calculated for each pair of chunks. The pair $\langle$ \textit{tall man}, \textit{man who is tall} $\rangle$ is a pair of paraphrases, and thus has a high alignment score (0.98). The pair $\langle$ \textit{tall man}, \textit{running along a road way} $\rangle$ has two unrelated phrases, and thus has a low alignment score(0.03). 
}
\label{graph-outline}
\end{figure*}
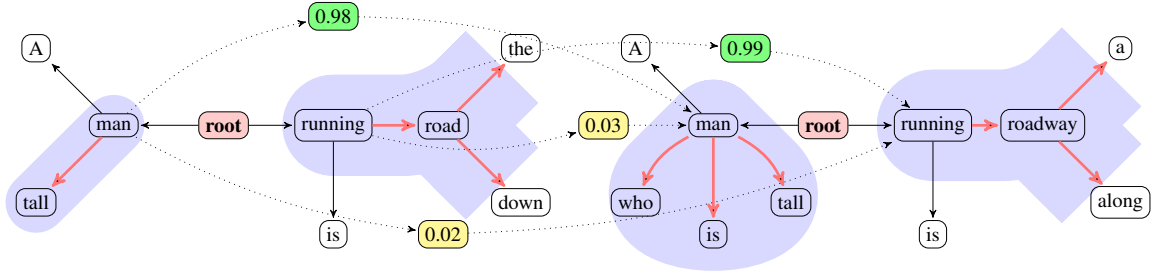

\subsection{Phrasal Monotonicity Inference}
Phrasal replacements are for phrase-level monotonicity inference. For example, with a polarized sentence
\textit{A $^\uparrow$ woman$^\uparrow$ who$^\uparrow$ is$^\uparrow$ beautiful$^\uparrow$ is$^\uparrow$ walking$^\uparrow$ in$^\uparrow$ the$^\uparrow$ rain$^=$}, the monotone mark $^\uparrow$ on \textit{woman} allows an upward inference: \textit{woman} $\sqsupseteq$ \textit{woman who is beautiful}, in which the relative clause \textit{who is beautiful} is deleted.   
The system follows a set of phrasal monotonicity inference rules. For upward monotonicity inference, modifiers of a word are deleted. For downward monotonicity inference, modifiers are inserted to a word. The algorithm traverses down a polarized UD parse tree, deletes the modifier sub-tree if a node is monotone ($\uparrow$), and inserts a new sub-tree if a node is antitone ($\downarrow$). To insert new modifiers, the algorithm extracts a list of potential modifiers associated to a node from a modifier dictionary. The modifier dictionary is derived from the hypothesis and contains word-modifier pairs for each dependency relation. Below is an example of a modifier dictionary from \textit{There are no beautiful flowers that open at night}:
\begin{itemize}
    \small 
    \item \textbf{obl}: [{head: \textit{open}, mod: \textit{at night}}]
    \item \textbf{amod}: [{head: \textit{flowers}, mod: \textit{beautiful}}]
    \item \textbf{acl:relcl}: [{head: \textit{flowers}, mod: \textit{that open at night}}]
\end{itemize}

\subsection{Syntactic Variation Inference}
We categorize linguistic changes between a premise and a hypothesis that cannot be inferred from monotonicity information as \emph{syntactic variations}. For example, a change from \textit{red rose} to \textit{a rose which is red} is a syntactic variation. Many logical systems rely on handcrafted rules and manual transformation to enable the system 
to use syntactic variations. However, without accurate alignments between the two sentences, these methods are not robust enough, and thus are difficult to scale up for wide-coverage input. 

Recent development of pretrained transformer-based language models are showing state-of-art performance on multiple benchmarks for Natural Language Understanding (NLU) including the task for paraphrase detection \cite{devlin-etal-2019-bert, Lan2020ALBERT:, liu2020roberta}
exemplify phrasal knowledge of syntactic variation. We propose a method that incorporates transformer-based language models to robustly handle syntactic variations. Our method first uses a sentence chunker to decompose both the premise and the hypothesis into chunks of phrases and then forms a Cartesian product of chunk pairs. For each pair, we use a transformer model to calculate the likelihood of a pair of chunks being a pair of paraphrases.  

\subsubsection{Sequence Chunking}
 To obtain phrase-level chunks from a sentence, we build a sequence chunker to extract chunks from a sentence using its universal dependency information. Instead of splitting a sentence into chunks, our chunker composes word tokens recursively to form meaningful chunks. First, we construct a sentence representation graph of a premise from the controller. Recall that a sentence representation graph is defined as $\mathrm{G} = \langle \mathcal{V}, \mathcal{E} \rangle$, where $\mathcal{V} = \mathcal{V}_{m} \cup \mathcal{V}_{c}$ is the set of modifiers ($\mathcal{V}_{m}$) and content words ($\mathcal{V}_{c}$), and $\mathcal{E}$ is the set of directed edges. To generate the chunk for a content word in $\mathcal{V}_{c}$, we arrange its modifiers, which are nodes it points to, together with the content word by their word orders in the original sentence to form a word chain. Modifiers that make the chain disconnected are discarded because they are not close enough to be part of the chunk. For instance, the chunk from the verb \textit{eats} in the sentence \textit{A person eats the food carefully} would not contain its modifier \textit{carefully} because they are separated by the object \textit{the food}. If the sentence is stated as \textit{A person carefully eats the food}, \textit{carefully} now is next to \textit{eat} and it would be included in the chunk of the verb \textit{eat}. To obtain chunks for a sentence, we iterate through each main component node, which is a node for subject, verb, or object, in the sentence's graph representation and construct verb phrases by combining verbs' chunks with their paired objects' chunks. There are cases when a word modifies other words and gets modified in the same time. They often occur when a chunk serves as a modifier. For example, in \textit{The woman in a pink dress is dancing}, the phrase \textit{in a pink dress} modifies \textit{woman} whereas \textit{dress} is modified by \textit{in}, \textit{a} and \textit{pink}. Then edges from \textit{dress} to \textit{in}, \textit{a}, \textit{pink} with the edge from \textit{woman} to \textit{dress} can be drawn. Chunks \textit{in a pink dress} and \textit{the woman in a pink dress} will be generated for \textit{dress} and \textit{woman} respectively.
 
 \subsubsection{Monolingual Phrase Alignment}
After the chunker outputs a set of chunks from a generated sentence and from the hypothesis, the system selects chunk pairs that are aligned by computing an alignment score for each pair of chunks. Formally, we define $\mathcal{C}_s$ as the set of chunks from a generated sentence and $\mathcal{C}_h$ as the set of chunks from the hypothesis. We build the Cartesian product from $\mathcal{C}_s$ and $\mathcal{C}_h$, denoted  $\mathcal{C}_s \times \mathcal{C}_h$. For each chunk pair ($\mathrm{c}_{si}$, $\mathrm{c}_{hj}) \in \mathcal{C}_s \times \mathcal{C}_h$, we compute an alignment score
$\boldsymbol{\alpha}$: 
\begin{align*}
    \yy_{\langle \mathbf{c_{si}}, \mathbf{c_{hi}} \rangle} &= \mathrm{ALBERT}.\mathrm{forward}(\langle \mathbf{c_{si}}, \mathbf{c_{hi}} \rangle) \\
    \boldsymbol{\alpha}_{\langle \mathbf{c_{si}}, \mathbf{c_{hi}} \rangle} &= \mathrm{p}(\mathbf{c_{si}} \mid \mathbf{c_{hj}}) \\
    \boldsymbol{\alpha}_{\langle \mathbf{c_{si}}, \mathbf{c_{hi}} \rangle} &= \frac{\exp^{\yy_{\langle \mathbf{c_{si}}, \mathbf{c_{hi}} \rangle_0}}}{\sum_{j=1}^2 \exp^{\yy_{\langle \mathbf{c_{si}}, \mathbf{c_{hi}} \rangle_ j}}}
\end{align*}
If $\boldsymbol{\alpha} > 0.85$, the system records this pair of phrases as a pair of syntactic variation. To calculate the alignment score, we use an ALBERT \cite{Lan2020ALBERT:} model for the paraphrase detection task, fine tuned on the Microsoft Research Paraphrase Corpus \cite{dolan-brockett-2005-automatically}. We first pass the chunk pair to ALBERT to obtain the logits. Then we apply a softmax function to the logits to get the final probability. A full demonstration of the alignment between chunks is shown in Figure \ref{graph-outline}.

\section{Data}
\subsection{The SICK Dataset}
The SICK \cite{marelli-etal-2014-sick} dataset is an English benchmark that provides in-depth evaluation for compositional distribution models. There are 10,000 English sentence pairs exhibiting a variety of lexical, syntactic, and semantic phenomena. Each sentence pair is annotated as Entailment, Contradiction, or Neutral. we use the 4,927 test problems for evaluation.

\subsection{The MED Dataset}
The Monotonicity Entailment Dataset (MED), is a challenge dataset designed to examine a model's ability to conduct monotonicity inference \citep{yanaka-etal-2019-neural}. There are 5382 sentence pairs in MED, where 1820 pairs are upward inference problems, 3270 pairs are downward inference problems, and 292 pairs are problems
with no monotonicity information. MED's problems cover a variety of linguistic phenomena, such as lexical knowledge, reverse, conjunction and disjunction, conditional, and negative polarity items.

\section{Evaluation}
\subsection{Experiment Setup}
For Universal Dependency parsing, we follow \citet{chengaoudep2mono}'s framework and use a parser from Stanford's natural language analysis package, Stanza \cite{qi-etal-2020-stanza}. In the parser, we use a neural parsing model pretrained on the UD English GUM corpus \cite{Zeldes2017} with 90.0 LAS \cite{zeman-etal-2018-conll} evaluation score. For Sentence-BERT, we selected the BERT-large model pre-trained on STS-B \cite{cer-etal-2017-semeval}. For ALBERT, we used textattack's ALBERT-base model pretrained on MRPC from transformers. For word alignment in the controller, we select \citet{rehurek_lrec}'s Gensim framework to calculate word similarity from pre-trained word embedding. 
We evaluated our model on the SICK and MED datasets using the standard NLI evaluation metrics of accuracy, precision, and recall. Additionally, we conducted two ablation tests focusing on analyzing the contributions of the monotonicity inference modules and the syntactic variation module. 

\begin{table}[t!]
\scalebox{0.85}{
\begin{tabular}{l|lll} \hline
	\textbf{Model} & \textbf{P} & \textbf{R} & \textbf{acc.} \\ \hline
	\multicolumn{4}{c}{\textbf{ML/DL-based systems}} \\\hline
	BERT (base, uncased) & 86.8 & 85.4 & 86.7 \\
	\citet{yin-schutze-2017-task} & -- & -- & 87.1 \\ 
	\citet{beltagy-etal-2016-representing} & -- & -- & 85.1 \\
	\hline
	\multicolumn{4}{c}{\textbf{Logic-based systems}} \\\hline
	\citet{abzianidze-2017-langpro} & 98.0 & 58.1& 81.4 \\
	\citet{martinez-gomez-etal-2017-demand} & 97.0 & 63.6 & 83.1 \\
	\citet{yanaka-etal-2018-acquisition} & 84.2 & 77.3 & 84.3 \\
	\citet{hu-etal-2020-monalog} & 83.8 & 70.7 & 77.2 \\
    \citet{abzianidze-2020-learning} & 94.3 & 67.9 & 84.4 \\
    \hline
    \multicolumn{4}{c}{\textbf{Hybrid System}} \\ \hline
    \citet{hu-etal-2020-monalog}+BERT &  83.2 &  85.5 & 85.4 \\
    \citet{kalouli-etal-2020-hy} & -- & -- & 86.5 \\
	\hline
	\multicolumn{4}{c}{\textbf{Our System}} \\ \hline
	NeuralLog (full system) & 88.0 & \textbf{87.6} & \textbf{90.3} \\
	$\,\,\,\,\,\,- \,\,$ALBERT-SV & 68.9 & 79.3 & 71.4 \\
    $\,\,\,\,\,\,- \,\,$Monotonicity & 74.5 & 75.1 & 74.7 \\
    \hline
\end{tabular}
}
\caption{Performance on the SICK test set  \label{tab:sick}}
\end{table}

\subsection{Results}
\paragraph{SICK}
Table \ref{tab:sick} shows the experiment results tested on SICK. We compared our performance to several logic-based systems as well as two deep learning based models. As the evaluation results show, our model achieves the state-of-art performance on the SICK dataset. The best logic-based model is \citet{abzianidze-2020-learning} with 84.4 percent accuracy. The best DL-based model is \citet{yin-schutze-2017-task} with 87.1 percent accuracy. Our system outperforms both of them with 90.3 percent accuracy. Compare to \citet{hu-etal-2020-monalog} + BERT, which also explores a way of combining logic-based methods and deep learning based methods, our system shows higher accuracy with a 4.92 percentage point increase. In addition, our system's accuracy has a 3.8 percentage point increase than another hybrid system, Hy-NLI \cite{kalouli-etal-2020-hy}. The good performance proves that our framework for joint logic and neural reasoning can achieve state-of-art performance on inference and outperforms existing systems. 

\paragraph{Ablation Test} In addition to the standard evaluation on SICK, we conducted two ablation tests. The results are included in Table \ref{tab:sick}. First, we remove the syntactic variation module that uses neural network for alignment ($-$ALBERT-SV). As the table shows, the accuracy drops 18.9 percentage points. This large drop in accuracy indicates that the syntactic variation module plays a major part in our overall inference process. The result also proves our hypothesis that deep learning methods for inference can improve the performance of traditional logic-based systems significantly. Secondly, when we remove the monotonicity-based inference modules ($-$Monotonicity), the accuracy shows another large decrease in accuracy, with  a 15.6 percentage point drop. This result demonstrates the important contribution of the logic-based inference modules toward the overall state-of-the-art performance. Compared to the previous ablation test which removes the neural network based syntactic variation module, the accuracy does not change much (only 3.3 differences). This similar performance indicates that neural network inference in our system alone cannot achieve state-of-art performance on the SICK dataset, and additional guidance and constrains from the logic-based methods are essential parts of our framework. Overall, we believe that the results reveal that both modules, logic and neural, contribute equally to the final performance and are both important parts that are unmovable.    

\begin{table}[t!]
\scalebox{0.85}{
\centering
\begin{tabular}{l|lll}
\hline \textbf{Model} & \textbf{Up} & \textbf{Down}& \textbf{All} \\ \hline
DeComp \citep{parikh-etal-2016-decomposable}& 71.1 & 45.2 &51.4\\
ESIM \citep{chen-etal-2017-enhanced} & 66.1 & 42.1 &53.8\\
BERT \citep{devlin-etal-2019-bert} & 82.7 & 22.8 &44.7 \\
BERT+ \citep{yanaka-etal-2019-neural} & 76.0 & 70.3 &71.6 \\
NeuralLog (ours) & \textbf{91.4} & \textbf{93.9} &\textbf{93.4} \\
\hline
\end{tabular}
}
\caption{\label{tab:med} Results comparing  model compared to state-of-art NLI models evaluated on MED. \textbf{Up}, \textbf{Down}, and \textbf{All} stand for the accuracy on upward inference, downward inference, and the overall dataset.}
\end{table}

\paragraph{MED}
Table \ref{tab:med} shows the experimental results tested on MED. We compared to multiple deep learning based baselines. Here, DeComp and ESIM are trained on SNLI and BERT is fine-tuned with MultiNLI. The BERT+ model is a BERT model fine-tuned on a combined training data with the HELP dataset, \cite{yanaka-etal-2019-help}, a set of  augmentations for monotonicity reasoning, and the MultiNLI training set. Both models were tested in \citet{yanaka-etal-2019-neural}. Overall, our system (NeuralLog) outperforms  all DL-based baselines in terms of accuracy, by a significant amount. Compared to BERT+, our system performs better both on upward (+15.4) and downward (+23.6) inference, and shows significant higher accuracy overall (+21.8). The good performance on MED validates our system's ability on accurate and robust monotonicity-based inference. 
 
\subsection{Error Analysis} 
For entailment, a large amount of inference errors are due to an incorrect dependency parse trees from the parser. For example, P: \textit{A black, red, white and pink dress is being worn by a woman}, H: \textit{A dress, which is black, red, white and pink is being worn by a woman}, has long conjunctions that cause the parser to produce two separate trees from the same sentence. Secondly, a lack of sufficient background knowledge causes the system to fail to make
inferences which would be needed to obtain a correct label. 
For example, P: \textit{One man is doing a bicycle trick in midair}, H: \textit{The cyclist is performing a trick in the air} requires the system to know that \textit{a man doing a bicycle trick} is a \textit{cyclist}. This kind of knowledge can only be injected to the system either by handcrafting rules or by extracting it from the training data. 
For contradiction, our analysis reveals inconsistencies in the SICK dataset. We account for multiple sentence pairs that have the same syntactic and semantic structures, but are labeled differently. For example, P: \textit{A man is folding a tortilla}, H: \textit{A man is unfolding a tortilla} has gold-label \textbf{Neutral} while P: \textit{A man is playing a guitar}, H: \textit{A man is not playing a guitar} has gold-label \textbf{Contradiction}. These two pair of sentences clearly have similar structures but have inconsistent gold-labels. Both gold-labels would be reasonable depending on whether the two subjects refer to the same entity. 
 
\section{Conclusion and Future Work}
In this paper, we presented a framework to combine logic-based inference with deep-learning based inference for improved Natural Language Inference performance. The main method is using a search engine and an alignment based controller to dispatch the two inference methods (logic and deep-learning) to their area of expertise. This way, logic-based modules can solve inference that requires logical rules and deep-learning based modules can solve inferences that contain syntactic variations which are easier for neural networks. Our system uses a beam search algorithm and three inference modules (lexical, phrasal, and syntactic variation) to find an optimal path that can transform a premise to a hypothesis. Our system handles syntactic variations in natural sentences using the neural network on phrase chunks, and our system determines contradictions by searching for contradiction signatures (evidence for contradiction). Evaluations on SICK and MED show that our proposed framework for joint logical and neural reasoning can achieve state-of-art accuracy on these datasets. Our experiments on ablation tests show that
neither logic nor neural reasoning alone fully solve Natural Language Inference, but a joint operation between them can bring improved performance. 

For future work, one plan
is to extend our system with more logic inference methods such as those using dynamic semantics \cite{haruta-etal-2020-combining} and more neural inference methods such as those for commonsense reasoning \cite{levine-etal-2020-sensebert}. We also plan to implement a learning method that allows the system to learn from mistakes on a training dataset and automatically expand or correct its rules and knowledge bases, which is similar to \citet{abzianidze-2020-learning}'s work.   

\section*{Acknowledgements}
We thank the anonymous reviewers for their insightful comments. We also thank Dr. Michael Wollowski from Rose-hulman Institute of Technology for his helpful feedback on this paper.

\bibliography{anthology,acl2021}
\bibliographystyle{acl_natbib}

\end{document}